\documentclass[11pt]{article}

\usepackage[utf8]{inputenc}
\usepackage{graphicx}
\usepackage{hyperref}
\usepackage{amsmath}
\usepackage{amsfonts}
\usepackage{amssymb}
\usepackage{geometry}
\title{A Modular Cognitive Architecture for Assisted Reasoning:\\
The Nemosine Framework}

\author{
  Edervaldo Jos\'e de Souza Melo \\
  \texttt{edersouzamelo@gmail.com} \\
  ORCID: \href{https://orcid.org/0009-0003-6835-135X}{0009-0003-6835-135X}
}

\date{December 2025}

\begin{document}

\maketitle

\begin{abstract}
This paper presents Nemosine Nous, a modular cognitive architecture designed to support assisted reasoning, mental organization, and metacognitive processes through a structured set of functional components. The framework is composed of coordinated modules that operate as cognitive agents, enabling task decomposition, perspective shifting, and symbolic processing. The development followed a design-science methodology combining theoretical analysis, structural modeling, and systematic documentation. An exploratory evaluation with 21 participants indicated high perceived clarity, intuitive understanding of the modular divisions, and usefulness for organizing thoughts and expanding analytical viewpoints. The proposed architecture contributes to the field of human–AI cognitive frameworks by offering an operational, implementation-oriented model suitable for individual use and future computational integration.

\end{abstract}

\section{Introduction}
Contemporary human–AI interaction increasingly requires cognitive support systems capable of organizing information, structuring reasoning processes, and reducing mental load during analysis and decision-making tasks. While existing tools—such as productivity applications, LLM-based assistants, and decision-support dashboards—offer partial assistance, they lack an integrated cognitive architecture capable of guiding users through systematic thought processes. Current solutions tend to emphasize surface-level execution (e.g., task lists, textual answers, or isolated suggestions) rather than providing a coherent internal structure for reasoning.

Research in cognitive science, distributed cognition, and metacognitive scaffolding suggests that external cognitive systems can enhance human reasoning \cite{hutchins1995,flavell1979} when they provide modular organization, perspective shifting, and structured workflows. However, few practical frameworks translate these theoretical principles \cite{fodor1983,norman1993} into operational, user-oriented architectures that support real-world cognitive tasks in a repeatable, interpretable, and customizable way.

Nemosine was developed to address this gap. It proposes a modular, function-driven cognitive architecture designed to assist individuals in organizing thoughts, clarifying analytical pathways, and sustaining metacognitive reflection. Rather than acting as a traditional assistant, Nemosine structures cognition itself through coordinated functional modules that operate as cognitive agents, enabling users to navigate problems using explicit, reproducible mental steps.

This paper introduces the architecture, describes its design-science development process, and reports an exploratory evaluation of its perceived clarity, usability, and cognitive benefits. The goal is to provide a technically grounded foundation for future implementations of Nemosine as a computational framework for assisted reasoning.

A Portuguese-language version of this work is also available as a preprint in SciELO Preprints 
(\href{https://preprints.scielo.org/index.php/scielo/preprint/view/14341}{link})
and in the dedicated Zenodo repository for the Nemosine system 
(\href{https://zenodo.org/communities/sistema-nemosine}{link}).

\section{Background}
Research on cognitive support systems has expanded across several domains, including cognitive science, human–computer interaction, and distributed cognition. Early work in cognitive psychology emphasized internal mental processes such as attention, working memory, and metacognition, highlighting how individuals monitor and regulate their own thinking. Subsequent models of distributed cognition \cite{maturana1998} demonstrated that cognitive processes frequently extend beyond the individual, relying on artifacts, tools, and external structures to support reasoning \cite{clark1998} and problem-solving.

In parallel, human–computer interaction studies have shown that well-designed computational systems can enhance cognitive performance when they provide structure, reduce cognitive load, or scaffold complex analytical tasks. Frameworks such as cognitive task analysis, sensemaking workflows, and metacognitive prompts illustrate how external systems can shape internal reasoning by organizing information and guiding decision steps.

Despite these contributions, existing tools rarely offer a unified cognitive architecture that is both modular and operationally oriented. Many systems provide isolated functions—such as task reminders, note-taking environments, LLM-generated suggestions, or decision-support dashboards—but lack an integrated framework capable of structuring reasoning processes in a consistent, interpretable way. As a result, users often rely on fragmented strategies, switching between multiple tools without a coherent mental model that binds them together.

This gap indicates a need for architectures that combine principles of distributed cognition, metacognitive scaffolding, and modular system design into a single, user-centered framework. Such architectures should enable individuals to externalize reasoning steps, shift perspectives systematically, and navigate analytical tasks using reproducible cognitive operations. These principles form the theoretical foundation for the development of Nemosine.

\section{The Nemosine Cognitive Architecture}
Nemosine is a modular cognitive architecture designed to externalize and structure human reasoning using a coordinated set of functional components. Each component operates as a cognitive agent responsible for a specific class of mental operations, and the architecture organizes these agents into a coherent system that supports assisted reasoning, perspective shifting, and metacognitive reflection. The model is implementation-oriented: its structure is defined such that it can be mapped to computational or hybrid human–AI systems in future research.

\subsection{Design Principles}

The architecture is based on four foundational principles:

\textbf{(1) Modularity.} Cognitive processes are decomposed into distinct functional units, each responsible for a specific type of operation such as analysis, evaluation, memory organization, emotional regulation, or strategic formulation.

\textbf{(2) Functional Coordination.} Modules do not operate in isolation; they follow predefined interaction rules that enable them to support one another. This coordination forms a cognitive workflow that guides users through structured reasoning steps.

\textbf{(3) Metacognitive Support.} Several modules are designed to monitor cognitive states, clarify goals, and maintain coherence of thought, allowing users to reflect on their own reasoning while performing it.

\textbf{(4) Interpretability.} The architecture prioritizes transparency by making cognitive steps explicit and reproducible, reducing reliance on opaque or intuitive processes. This feature also supports potential computational implementations.

\subsection{Modular Structure}

Nemosine is composed of multiple cognitive agents organized into functional categories, including analytical processing, decision support, emotional modulation, memory structuring, and perspective management. Each agent represents a standardized cognitive operation that can be invoked, sequenced, or combined depending on the user’s task.

Although the architecture allows a large number of agents, its core structure is defined by a stable set of high-level modules that shape the overall cognitive workflow. These modules act as anchors for reasoning and metacognitive control, ensuring consistency across tasks.

\subsection{Cognitive Workflow}

The cognitive workflow follows a stepwise process:

\textbf{(1) Problem Framing.} The system begins by clarifying the user’s objective, constraints, and context.

\textbf{(2) Perspective Activation.} Relevant cognitive agents are activated to provide analytical, emotional, evaluative, or strategic viewpoints on the task.

\textbf{(3) Structured Processing.} Modules operate in sequence or in parallel to decompose the problem, evaluate alternatives, and generate structured reasoning paths.

\textbf{(4) Metacognitive Monitoring.} Dedicated modules track coherence, detect inconsistencies, and ensure that the reasoning remains aligned with the task goals.

\textbf{(5) Synthesis and Output.} The system integrates the outputs of active modules into a final structured representation, which can take the form of a decision pathway, written analysis, or cognitive summary.

\subsection{Implementation Orientation}

Although the present work focuses on the conceptual architecture, the modular structure is compatible with computational implementations. Each cognitive agent can be mapped onto:

- symbolic rule-based systems,  
- multi-agent reasoning frameworks,  
- LLM-assisted cognitive operators, or  
- hybrid human–AI decision-support interfaces.

This compatibility positions Nemosine as a candidate framework for the development of assistive cognitive systems capable of guiding structured reasoning in real-world contexts.

\section{Method}
The development of Nemosine followed a design-science methodology typically applied to the construction of conceptual and computational frameworks. The process consisted of four stages: theoretical analysis, structural modeling, systematic documentation, and exploratory evaluation.

\subsection{Theoretical Analysis}

The initial stage involved reviewing literature on cognition, distributed cognition, metacognition, and human–computer interaction to identify principles relevant to assisted reasoning. These principles informed the conceptual boundaries of the architecture and guided decisions about modularity, workflow, and functional roles.

\subsection{Structural Modeling}

In the second stage, the architecture was modeled as a set of coordinated cognitive agents, each representing a distinct class of mental operations. The model was refined iteratively to ensure internal coherence, functional complementarity, and compatibility with potential computational implementations. Diagrams, templates, and operational definitions were produced to standardize the structure of each module.

\subsection{Systematic Documentation}

The third stage focused on organizing the architecture into a formal, reproducible structure. Documentation included defining module interactions, workflow sequences, activation rules, and metacognitive control mechanisms. This stage ensured that the architecture could be interpreted consistently by users and could serve as a basis for future computational translation.

\subsection{Exploratory Evaluation Design}

The final stage consisted of an exploratory evaluation conducted with 21 voluntary participants, who interacted with the conceptual version of the architecture and provided assessments of clarity, modular understanding, and perceived usefulness. Data collection followed a structured questionnaire combining closed and open-ended items. Quantitative responses were summarized descriptively, and qualitative feedback was analyzed to identify patterns of perceived cognitive benefit.

The purpose of this evaluation was not to validate the architecture empirically, but to assess its initial interpretability and practical clarity for users, informing future refinements and guiding subsequent development work.

\section{Exploratory Evaluation}
An exploratory evaluation was conducted to assess the initial interpretability and perceived usefulness of the Nemosine architecture. The goal of this evaluation was not to empirically validate the framework, but to examine how users understood its modular structure and whether they perceived cognitive benefits when interacting with it.

\subsection{Participants}

Twenty-one voluntary participants completed the evaluation. They represented a heterogeneous group of adults with varied educational and professional backgrounds. Participation was anonymous and no identifying information was collected.

\subsection{Procedure}

Participants were presented with the conceptual version of the Nemosine architecture, including its modular structure, workflow, and examples of cognitive operations. They then completed a structured questionnaire containing closed questions (rated on Likert-type scales) and open-ended prompts inviting qualitative feedback.

The closed questions assessed three domains:
\begin{itemize}
    \item \textbf{Perceived clarity} of the architecture;
    \item \textbf{Intuitive understanding} of the modular divisions;
    \item \textbf{Perceived usefulness} for organizing thoughts or expanding analytical perspectives.
\end{itemize}

\subsection{Results}

Quantitative responses indicated high perceived clarity and coherence of the modular structure. Most participants reported that the division into functional modules was intuitive and helped them understand how the architecture organizes cognitive processes. Open-ended responses revealed recurring themes, including improved ability to structure reasoning, increased awareness of cognitive steps, and perceived assistance in organizing complex thoughts.

\begin{table}[h!]
\centering
\begin{tabular}{l l c}
\hline
\textbf{Item} & \textbf{Most frequent response} & \textbf{\%} \\
\hline
Perceived clarity & Clara & 42.9 \\
Understanding of the system & Sim & 90.5 \\
Modular division makes sense & Equilibrada & 66.7 \\
Perception of language & Moderada & 52.4 \\
Perceived complexity & Moderada & 52.4 \\
Would recommend the system & Sim & 85.7 \\
Helped organize thoughts & Sim & 81.0 \\
Expanded perspective & Sim & 76.2 \\
\hline
\end{tabular}
\caption{Summary of the most frequent participant responses for each evaluated variable (n = 21).}
\label{tab:resultsfreq}
\end{table}

Although exploratory in nature, the results suggest that users were able to interpret the architecture consistently and recognize potential cognitive benefits. These findings informed the refinement of the documentation and support the viability of developing future computational implementations based on the framework.

\section{Discussion}
The exploratory evaluation suggests that users were able to understand and apply the modular structure of Nemosine, perceiving it as a coherent framework for organizing reasoning processes. These results align with prior research on distributed cognition and metacognitive scaffolding, which indicates that external structures can enhance clarity and reduce cognitive load when they make reasoning steps explicit.

The architecture’s emphasis on modularity and functional coordination appears to support users in decomposing complex tasks and shifting perspectives systematically. Participants reported that the clearly defined modules helped them articulate thought processes that are often implicit or intuitive, which may contribute to improved self-monitoring and cognitive organization.

However, because the evaluation was exploratory and based on self-reported perceptions, the findings should be interpreted cautiously. The study does not provide empirical validation of cognitive performance improvements, nor does it assess long-term use or task-specific effectiveness. Future work should examine how the architecture performs in real-world contexts, whether users maintain consistent interaction patterns, and how the model can be integrated into computational systems.

Despite these limitations, the results indicate that Nemosine has potential as a structured cognitive support framework. Its modular design may offer a foundation for developing hybrid human–AI systems that guide users through systematic reasoning steps while maintaining transparency and interpretability.

\section{Conclusion}
This paper presented Nemosine Nous, a modular cognitive architecture designed to support assisted reasoning through structured cognitive operations and coordinated functional components. The framework integrates principles from distributed cognition, metacognitive scaffolding, and human–computer interaction to provide a clear, interpretable structure for organizing thought processes.

An exploratory evaluation with 21 participants indicated that users were able to understand the architecture and perceive potential cognitive benefits. Although these findings are preliminary and based on self-reported perceptions, they provide initial support for the clarity and coherence of the framework.

Future work should develop computational implementations of the architecture, investigate task-specific performance effects, and examine long-term patterns of user interaction. By advancing these directions, Nemosine may serve as a basis for hybrid human–AI systems capable of guiding structured reasoning in real-world contexts.

\bibliographystyle{unsrt}
\bibliography{references}

\end{document}